\pdfoutput=1

\documentclass[11pt]{article}

\usepackage[]{EMNLP2022}

\usepackage{times}
\usepackage{latexsym}

\usepackage[T1]{fontenc}

\usepackage[utf8]{inputenc}

\usepackage{microtype}

\usepackage{inconsolata}

\usepackage{graphicx}
\usepackage{cleveref}
\usepackage{booktabs}

\usepackage[normalem]{ulem}

\usepackage{multirow}
\usepackage{url}

\usepackage{amsfonts}

\usepackage{xspace}

\newcommand{\draft}[1]{{}}

\newcommand{\ACE}[0]{ACE\xspace}
\newcommand{\ASR}[0]{ASR\xspace}
\newcommand{\AED}[0]{AED\xspace}
\newcommand{\AEDFULL}[0]{ASR Error Detection (AED) \xspace}

\newcommand{\LibriSpeech}[0]{LibriSpeech\xspace}

\newcommand{\clean}[0]{\emph{clean}\xspace}
\newcommand{\other}[0]{\emph{other}\xspace}

\newcommand{\ERR}{\textsc{Error}\xspace}
\newcommand{\NOTERR}{\textsc{NotError}\xspace}

\newcommand{\CONFIDENCE}{\textsc{C-O}\xspace}
\newcommand{\MLM}{\textsc{BERT-MLM}\xspace}
\newcommand{\BERT}{\textsc{BERT}\xspace}
\newcommand{\BERTorC}{\textsc{BERT | C}\xspace}
\newcommand{\BERTandC}{\textsc{BERT \& C}\xspace}
\newcommand{\BERTC}{\textsc{BERT$_C$}\xspace}
\newcommand{\MODEL}{\textsc{RED-ACE}\xspace}

\newcommand{\GED}{\textsc{GED}\xspace}

\newcommand{\ace}[0]{\includegraphics[width=.02\textwidth]{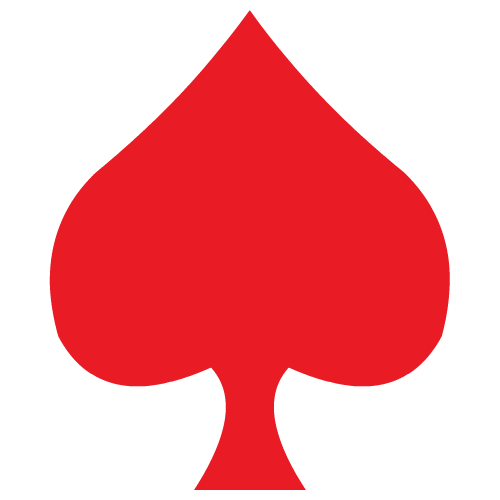}}
\newcommand{\bigace}[0]{\includegraphics[width=.025\textwidth]{figures/ace.png}}
\newcommand{\starr}{$^*$}

\newcommand{\video}[0]{\emph{video}\xspace}
\newcommand{\default}[0]{\emph{default}\xspace}

\title{\bigace \ \MODEL: Robust Error Detection for ASR using Confidence Embeddings}

\author{Zorik Gekhman\thanks{\ \ \  Equal contribution.}\hspace{0.5cm}Dina Zverinski\protect\footnotemark[1]\hspace{0.5cm}Jonathan Mallinson\hspace{0.5cm}Genady Beryozkin\\
Google Research\\
  {\tt \{zorik,zverinski,jonmall,genady\}@google.com}}
  
\begin{document}
\maketitle
    \begin{abstract}

\AEDFULL models aim to post-process the output of Automatic Speech Recognition (ASR) systems, in order to detect transcription errors. 
Modern approaches usually use text-based input, comprised solely of the ASR transcription hypothesis, disregarding additional signals from the ASR model. 
Instead, we utilize the ASR system's word-level confidence scores for improving \AED performance.
Specifically, we add an ASR Confidence Embedding (\ACE) layer to the \AED model's encoder, allowing us to jointly encode the confidence scores and the transcribed text into a contextualized representation.
Our experiments show the benefits of ASR confidence scores for AED, their complementary effect over the textual signal, as well as the effectiveness and robustness of \ACE for combining these signals.
To foster further research, we publish a novel \AED dataset consisting of ASR outputs on the LibriSpeech corpus with annotated transcription errors.\footnote{Our code and data are available at \url{https://github.com/google-research/google-research/tree/master/red-ace}.}

\end{abstract}
    \section{Introduction}
\label{sec:intro}

Automatic Speech Recognition (ASR) systems transcribe audio signals, consisting of speech, into text.
While state-of-the-art ASR systems reached high transcription quality, training them requires large amounts of data and compute resources.
Fortunately, many high performing systems are available as off-the-shelf 
cloud services.
However, a performance drop can be observed when applying them to specific domains or accents \cite{BLACKBOX_ADAPTATION, PREV_SEQ2SEQ_ERROR_CORRECTION_1}, or when transcribing noisy audio.
Moreover, cloud services usually expose the ASR models as a black box, 
making it impossible to further fine-tune them.

\AEDFULL models are designed to post-process the ASR output, in order to detect transcription errors and avoid their propagation to downstream tasks \cite{ERROR_CORRECTION_REVIEW}.
\AED models are widely used in interactive systems, to engage the user to resolve the detected errors.
For example, AED systems can be found in \emph{Google Docs Voice Typing}, where low confidence words are underlined, making it easier for users to spot errors and take actions to correct them.

\begin{figure}[t]
 \centering
    \includegraphics[width=\columnwidth]{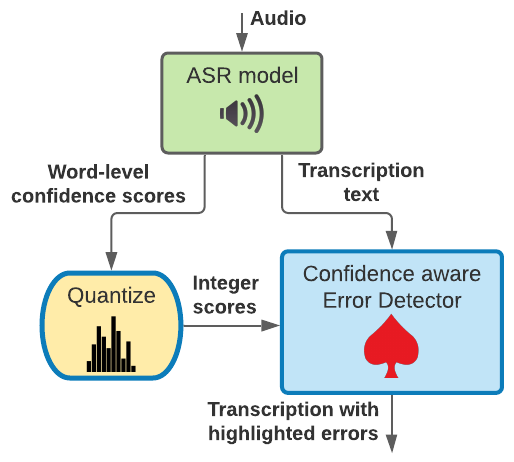}
      \vspace{-0.5cm}
    \caption{
    Our \AED pipeline. The confidence scores are quantized and jointly encoded with the transcription text into a contextualized representation.
    }
    \label{fig:diagram} 
    \vspace{-0.5cm}
\end{figure}

Modern NLP models usually build upon the Transformer architecture \cite{Transformer}. However, no Transformer-based AED models have been proposed yet. Recently, the Transformer has been applied to ASR \emph{error correction} \cite{PREV_SEQ2SEQ_ERROR_CORRECTION_1, PREV_SEQ2SEQ_ERROR_CORRECTION_2, FASTCORRECT2, FASTCORRECT}, another ASR post-processing task. These models use only the transcription hypothesis text as input and discard other signals from the ASR model.
However, earlier work on \AED (not Transformer-based) has shown the benefits of such signals \cite{ERROR_DETECTION_1, ERROR_DETECTION_2, ERROR_DETECTION_3} and specifically the benefits of ASR word-level confidence scores \cite{ERROR_DETECTION_WITH_CONFIDENCE}, which are often provided in addition to the transcribed text 
\cite{CONFIDENCE_SCORES_SURVEY, CONFIDENCE_SCORE_RESEARCH_1, CONFIDENCE_SCORE_RESEARCH_2}.

In this work we focus exclusively on \AED and propose a natural way to embed the ASR confidence scores into the Transformer architecture. We introduce \mbox{\ace \ \MODEL}, a modified Transformer encoder with an additional embedding layer, that jointly encodes the textual input and the word-level confidence scores into a contextualized representation (\Cref{fig:method}). 
Our \AED pipeline first quantizes the confidence scores into integers and then feeds the quantized scores with the transcribed text into the modified Transformer encoder (\Cref{fig:diagram}).
Our experiments demonstrate the effectiveness of \MODEL in improving \AED performance. In addition, we demonstrate the robustness of \MODEL to changes in the transcribed audio quality.
Finally, we release a novel dataset that can be used to train and evaluate \AED models.

    \section{\ace \ \MODEL}
\label{sec:method}
Following recent trends in NLP, we use a pre-trained Transformer-based language model, leveraging its rich language representation.
\MODEL is based on a pre-trained BERT \cite{BERT}, adapted to be confidence-aware and further fine-tuned for sequence tagging.
Concretely, our \AED model is a binary sequence tagger that given the ASR output, consisting of the transcription hypothesis words and their corresponding word-level confidence scores, predicts an \ERR or \mbox{\NOTERR} tag for each input token.\footnote{We discuss words to tokens conversion in \S \ref{impl:details}.}

Our \AED pipeline is illustrated in \Cref{fig:diagram}.
First, we quantize the floating-point confidence scores into integers using a binning algorithm.\footnote{Typical confidence scores range between 0.0 to 1.0. We perform experiments with simple equal-width binning and quantile-based discretization (equal-sized buckets), as well as different bin numbers. More details in \S \ref{impl:details}.}
Next, the quantized scores and the transcription text are fed into a confidence-aware BERT tagger.

\begin{figure}[t]
 \centering
  \vspace{-0.1cm}
 \includegraphics[width=\columnwidth]{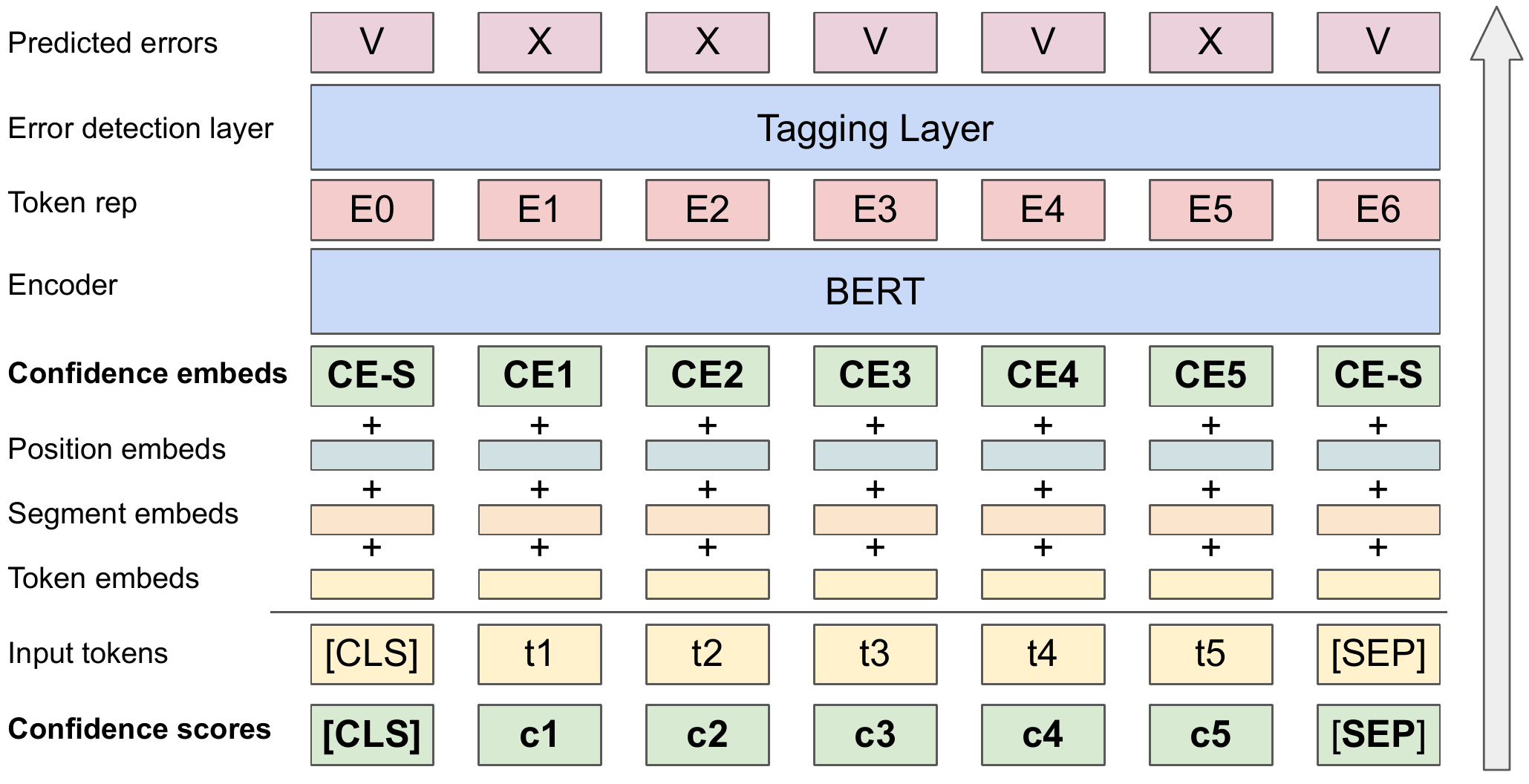}   
    \caption{Our confidence-aware AED model. We use a BERT-based tagger with modifications colored in green. An additional embedding layer is added to represent the embedding of the quantized confidence scores.}
    \label{fig:method}    
\end{figure}

In BERT, each input token has 3 embeddings: token, segment and position.\footnote{We refer the reader to \citet{BERT} for more details.}
To adapt BERT to be confidence-aware, we implement an additional dedicated embedding layer, indicating the confidence bin that the input token belongs to. We construct a learned confidence embedding lookup matrix $M \in \mathbb{R}^{B\times H}$, where $B$ is the number of confidence bins and $H$ is BERT's embedding vector's size. For a given token, its input representation is constructed by summing the corresponding BERT's embeddings with its confidence embedding (\Cref{fig:method}).
This allows the model to learn a dedicated dense representation vector for each confidence bin, as well as naturally combine it with the final contextualized representation of each input token.

\begin{table}[t]
\small
\scalebox{0.8}{
\centering
\setlength{\tabcolsep}{4pt}
\begin{tabular}{llcrrr}
\toprule
\multicolumn{1}{c}{ASR Model} & \multicolumn{1}{c}{Pool} & \multicolumn{1}{c}{Split} & \multicolumn{1}{c}{\# Examples} & \multicolumn{1}{c}{\# Words} & \multicolumn{1}{c}{\# Errors} \\
\midrule
\multirow{6}{*}{\emph{default}} & \multirow{3}{*}{\clean} & Train     &  103,895  &   3,574,027 & 357,145 (10.0\%)      \\
& & Dev       &   2,697&   54,062 & 5,111 (9.5\%)       \\
& & Test      &   2,615 &   52,235 & 4,934 (9.4\%)        \\
\cmidrule{2-6}
& \multirow{3}{*}{\other}  & Train  &   146,550 &     4,650,779 & 770,553 (16.6\%)     \\
& & Dev  &   2,809 &     48,389 & 9,876 (20.4\%)     \\
& & Test &   2,925 &     50,730 & 10,317 (20.3\%)     \\
\midrule
\multirow{6}{*}{\emph{video}} & \multirow{3}{*}{\clean} & Train     &  104,013  &   3,589,136 & 210,324 (5.9\%)      \\
& & Dev       &   2,703&   54,357 & 3,109 (5.7\%)       \\
& & Test      &   2,620 &   52,557 & 2,963 (5.6\%)        \\
\cmidrule{2-6}
& \multirow{3}{*}{\other}  & Train  &   148,678 &     4,810,226 & 148,678 (7.9\%)     \\
& & Dev  &   2,809 &     50,983 & 5,901 (11.6\%)     \\
& & Test &   2,939 &     52,192 & 6,033 (11.6\%)     \\
\bottomrule
\end{tabular}
}
\caption{ 
\AED dataset statistics.
}
\label{tab:dataset-stats}
\vspace{-0.5cm}
\end{table}

    \section{Dataset Creation and Annotation}
\label{sec:dataset}
To train and evaluate \AED models, we generate a dataset with labeled transcription errors.

\paragraph{Labeling of ASR Errors.}
We decode audio data using an \ASR model and obtain the transcription hypothesis. Then, we align the hypothesis words with the reference (correct) transcription. 
Specifically, we find an edit path, between the hypothesis and the reference, with the minimum edit distance and obtain a sequence of edit operations (insertions, deletions and substitutions) that can be used to transform the hypothesis into the reference. 
Every incorrect hypothesis word (i.e needs to be deleted or substituted) is labeled as \ERR, the rest are labeled as \NOTERR.

\paragraph{Audio Data Source.}
We use the \LibriSpeech corpus \cite{LIBRISPEECH}, containing 1000 hours of transcribed English speech from audio books.\footnote{\url{https://www.openslr.org/12/}}
The corpus contains \emph{clean} and \emph{other} pools, where \emph{clean} is of higher recording quality.\footnote{We provide additional details about the corpus in \S \ref{appendix:published_dataset}.}

\paragraph{ASR Models.}
In this work we focus exclusively on a black-box setting, where the exact implementation details of the ASR and the confidence models are unknown. This setting is particularly relevant since many applications rely on strong performance of black-box ASR models which are exposed as cloud services.
We use Google Cloud Speech-to-Text API as our candidate ASR model.\footnote{\url{https://cloud.google.com/speech-to-text}} 
In our main experiments we select the \default ASR model.\footnote{\url{https://cloud.google.com/speech-to-text/docs/basics\#select-model}}
To ensure the generalization ability of RED-ACE, we repeat our main experiments using a different ASR model, in this case we choose the \video model.
Table \ref{tab:dataset-stats} presents the statistics of our dataset.
It is notable that the main model’s error rate (\default) is about twice as high as the additional model’s error rate (\video), which shows that (even though both models are part of the Google Cloud API) this additional ASR model is substantially different from the main ASR model we used.

\paragraph{Data Release.}
Since Google Cloud requires a paid subscription and since the underlying ASR models may change over time, we make our dataset publicly available.\footnote{Additional details about the dataset are provided in \S \ref{appendix:libriSpeech_corpus}.} This ensures full reproducibility of our results (in case the ASR models change) and makes it easier for researchers to train and evaluate AED models, removing the need to run inference on paid cloud-based ASR models or train dedicated models in order to transcribe audio.
    \section{Experimental Setup}
\label{sec:experimental-setup}

Our experiments examine the informativeness of the confidence scores as well as the effectiveness of RED-ACE in combining them with text. We provide extensive implementation details in \S \ref{impl:details}.

\begin{figure}[t]
 \centering
    \includegraphics[width=0.95\columnwidth]{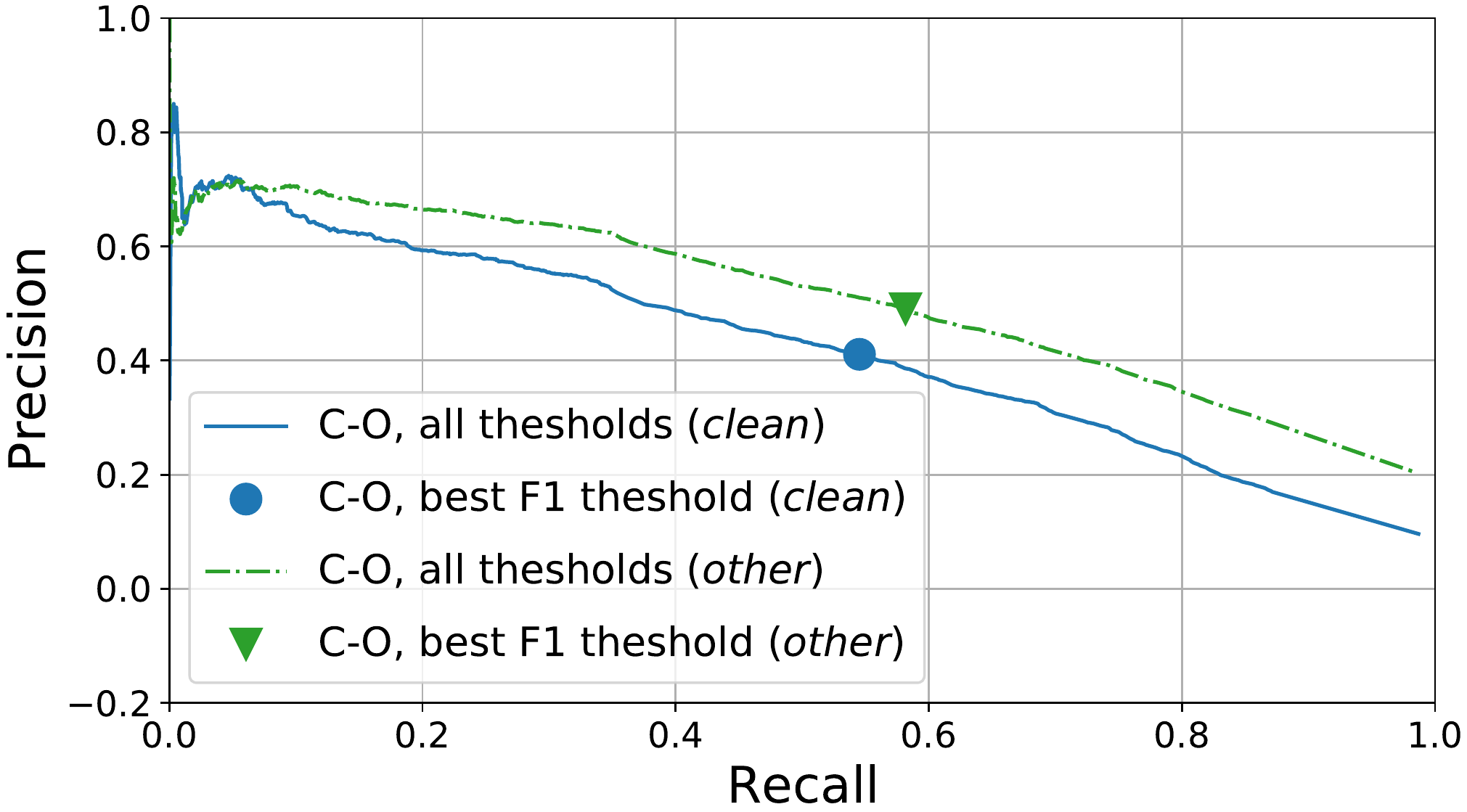}
    \vspace{-0.3cm}
    \caption{
    Threshold tuning process for the \CONFIDENCE baseline. Models are evaluated using different confidence scores thresholds and the threshold that yields the best F1 is chosen.
    A similar process is performed for \BERTandC and \BERTorC. For \MLM we tune the values for $k$. 
    }
    \label{fig:threshold_plot} 
\end{figure}

\begin{figure}[t]
 \centering
  \vspace{-0.1cm}
 \includegraphics[width=\columnwidth]{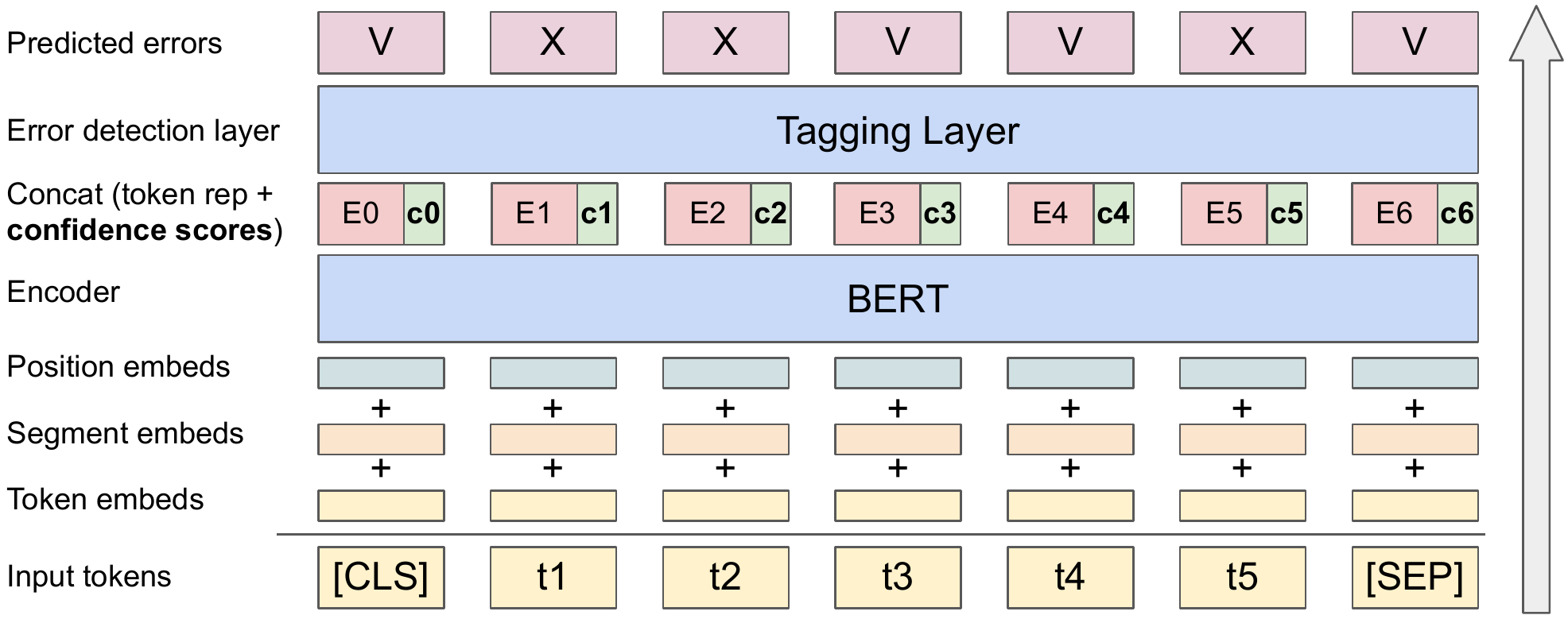}   
    \vspace{-0.5cm}
    \caption{The \BERTC baseline, which modifies the input to the tagger, unlike \MODEL which modifies BERT's embeddings. The value of the respective confidence score is appended to the final contextualized representation.}
    \label{fig:bertc}    
    \vspace{-0.5cm}
\end{figure}

\subsection{Baselines}
\label{sec:models}

\paragraph{\CONFIDENCE (Confidence Only)}
Uses the word-level scores from the ASR confidence model directly. Predicts \ERR if the token's confidence score is below a threshold.\footnote{We choose the confidence threshold or $k$ value (in case of \MLM) with the best F1 on the dev set (\Cref{fig:threshold_plot}).}

\paragraph{\MLM}
Masks out the input words one at a time and uses a pre-trained BERT \cite{BERT} as a Masked Language Model (MLM) in order to infill them. Predicts \ERR for input words that are not in the top $k$ BERT's suggestions.\footnotemark[9]

\paragraph{BERT} 
We fine-tune BERT \cite{BERT} for sequence tagging (only on text, \emph{without} adding \MODEL). 
As Transformers have not beeen applied for \AED yet,
we choose BERT as a pre-trained LM following \citet{BERT_GRAMMAR_ERROR_DETECTION}, who applied it for Grammatical Error Detection (\GED) and achieved the highest performance in the \textsc{NLPTEA-2020} Shared Task \cite{rao2020overview}.

\paragraph{BERT \& C}
Predicts \ERR if BERT predicts \ERR \textbf{and} confidence is below a threshold.\footnotemark[9]

\paragraph{BERT | C}
Predicts \ERR if BERT predicts \ERR \textbf{or} confidence is below a threshold.\footnotemark[9]

\paragraph{BERT$_C$}
We fine-tune \BERT \emph{jointly} with the confidence scores by concatenating the score value to the token's contextualized representation produced by BERT (directly before it is fed into the sequence tagger). 
BERT's last hidden layer dimension is increased by 1, and the corresponding value populated with the token's confidence score.
An illustration is provided in \Cref{fig:bertc}. 

\subsection{Evaluation}

\paragraph{Main Settings.}
In the main settings we train the models on the \clean and \other training sets and evaluate them \clean and \other test sets respectively.

\paragraph{Robustness Settings.}
A real-word \AED system should remain effective when the audio stems from different recording qualities.
Changes in recording quality, can affect the ASR model's errors distribution and thus can potentially reduce the effectiveness of the \AED model. 
As our dataset contains 2 pools with different recording quality (\Cref{tab:dataset-stats}), we can measure whether \MODEL's performance deteriorates when the audio quality of the training data changes.
To this end we define the \emph{robustness settings} (\Cref{tab:error-detection-robustness}), where we perform a cross-pools evaluation, evaluating models that were trained on \clean and \other training sets using the \other and the \clean test sets respectively.

\paragraph{Metric.}
We measure errors detection \emph{Precision (P)}, \emph{Recall (R)} and \emph{F1}. \emph{Recall} measures the percent of real errors that were detected, while \emph{Precision} measures the percent of the real errors out of all detected errors. We calculate the \emph{P} and  \emph{R} on the word-level. We also report span-level results for the main settings in \Cref{tab:span-level} in the appendix.

\begin{table}[t]
\small
\centering
\scalebox{0.83}{
\begin{tabular}{l|ccc|ccc}
\toprule
& \multicolumn{3}{c|}{   \clean} & \multicolumn{3}{c}{   \other} \\
  & R & P  & F1   & R &  P & F1 \\
\midrule
\CONFIDENCE    & 52.1 & 42.5 & 46.8 & 63.5 & 45.6 & 53.1  \\
\MLM       &    58.0   &    26.5   &    36.4   &    \textbf{72.7}   &    35.9   &    48.1     \\
\BERT       &    58.5          & 77.6 & 66.7              & 58.0          &  77.1             & 66.2  \\
\BERTandC    & 55.8    &  75.0 & 64.0    &  55.5 & 75.5           &  64.0 \\
\BERTorC   & \textbf{63.3}     &  68.1 &65.6    & 68.1 & 67.1           &  67.6 \\
\BERTC   & 51.7     &  78.9 & 66.3    & 58.1 & 78.8           &  66.9 \\
\midrule
\MODEL  &  61.1  & \ \ \textbf{81.9}\starr & \ \ \textbf{70.0}\starr     &  64.1 &  \ \ \textbf{79.9}\starr  & \ \ \textbf{71.1}\starr  \\
\midrule
F1 $\Delta$\%  & & & +4.9\% & & & +7.4\% \\
\bottomrule
\end{tabular}
}
\caption{
Main settings using the errors from the \default ASR model (see \Cref{tab:dataset-stats}).
R and P stands for Recall and Precision. F1 $\Delta$\% compares \MODEL to the BERT baseline.
Results with $*$ indicate a statistically significant improvement compared to the strongest baseline.
}
\label{tab:error-detection}
\vspace{-0.5cm}
\end{table}

    \section{Results} 
\label{sec:results}

\Cref{tab:error-detection} presents our main results, evaluating the models on the \emph{main settings} using errors from the main (\default) ASR model. \Cref{tab:error-detection-robustness} presents the results on the \emph{robustness settings}, also using errors from the main ASR model.

The low F1 of \mbox{\CONFIDENCE} suggest that the \ASR confidence has low effectiveness without textual signal. The low F1 of \MLM indicates that supervised training on real transcription errors is crucial. 

We next observe that \BERTandC performs worse than \BERT on all metrics. When comparing \mbox{\BERTorC} to \BERT we observe the expected increase in recall (\BERT's errors are a subset of the errors from \BERTorC) and a decrease in precision, F1 decreases on \clean and increases on \other. 
The results on \BERTC are particularly surprising. Similarly to \MODEL, \BERTC trains \BERT \emph{jointly} with the scores. However, unlike \MODEL, \BERTC performs worse than \BERT.
This demonstrates the effectiveness and importance of our modeling approach, that represents the scores using a learned dense embedding vectors.
As \MODEL is the only method that successfully combines the scores with text, we focus the rest of the analysis on comparing it to the text-based \BERT tagger.

In the \emph{main settings} (\Cref{tab:error-detection}), \MODEL consistently outperforms BERT on all evaluation metrics in both pools.
This demonstrates the usefulness of the confidence signal on top of the textual input, as well as the effectiveness of \MODEL in combining those signals.
\MODEL's F1 $\Delta$\% on \clean is lower than on \other.
This can be attributed to the fact that the error rate in \clean is twice lower than in \other (\Cref{tab:dataset-stats}), which means that the model is exposed to fewer errors during training.

\begin{table}[t]
\small
\centering
\scalebox{0.8}{
\begin{tabular}{l|ccc|ccc}
\toprule
& \multicolumn{3}{c|}{\other \ $\rightarrow$  \clean} & \multicolumn{3}{c}{\clean \ $\rightarrow$ \other} \\
  & R & P  & F1   & R &  P & F1 \\
\midrule
BERT    &    64.3           & 71.9 & 67.9              & 47.1          &  80.3             & 59.4 \\
\MODEL & \ \ \textbf{67.9}\starr  &   \ \ \textbf{77.0}\starr   & \ \ \textbf{72.2}\starr & \ \ \textbf{53.7}\starr     & \ \ \textbf{83.3}\starr &  \ \ \textbf{65.3}\starr  \\
\midrule
F1 $\Delta$\%  & & & +6.3\% & & & +9.9\% \\
\bottomrule
\end{tabular}
}
\caption{
Robustness settings with the \default ASR model (\Cref{tab:dataset-stats}). \other \ $\rightarrow$   \clean means train on \other and eval on \clean.
Format is similar to \Cref{tab:error-detection}.
}
\label{tab:error-detection-robustness}
\vspace{-0.5cm}
\end{table}

Finally, we analyze the \emph{robustness settings} from \Cref{tab:error-detection-robustness}. 
We first note that \MODEL outperforms BERT in both settings, indicating its robustness across different settings, and that it can remain effective with recording quality differences between train and test time. When observing the performance on the \clean test set, we observe that training \AED models on \other instead of \clean, leads to improvement in F1. This can be attributed to the higher error rate and larger number of training examples in \other (see \Cref{tab:dataset-stats}), which exposes the models to larger amount of errors during training. The F1 $\Delta$\% on \other \ $\rightarrow$  \clean (\Cref{tab:error-detection-robustness}) is comparable to \clean (\Cref{tab:error-detection}), with a statistically insignificant improvement. 
An opposite trend can be seen on the \other test set. The performance of models that were trained on \clean instead of \other deteriorates. Yet, \MODEL's relative performance drop is smaller than BERT's. 
\MODEL drops by $8.2\%$ (from $71.1$ to $65.3$) while BERT by $10.3\%$ (from $66.2$ to $59.4$). This is also demonstrated by the statistically significant increase in F1 $\Delta$\%, from $7.4\%$ in \other \ $\rightarrow$ \other to      $9.9\%$ in \clean \ $\rightarrow$ \other.
This serves as additional evidence for the robustness of \MODEL.
We also note that \clean \ $\rightarrow$ \other is the most challenging setting, with BERT's F1 significantly lower than the other 3 settings, meaning that \MODEL shows the largest improvement (F1 $\Delta$\%) in the hardest setting.

\paragraph{Generalization Across ASR Models.}
As discussed in \S \ref{sec:dataset}, to ensure that RED-ACE is applicable to not only one specific ASR model, we repeat our experiments using a different ASR model.
The results are presented in \Cref{tab:another_asr} and \Cref{tab:error-detection-robustness-video}. \MODEL outperforms BERT in all settings, with statistically significant F1 improvements, further highlighting \MODEL robustness.

\begin{table}[t]
\small
\centering
\scalebox{0.82}{
\begin{tabular}{l|ccc|ccc}
\toprule
& \multicolumn{3}{c|}{\clean} & \multicolumn{3}{c}{\other} \\
  & R & P  & F1   & R &  P & F1 \\
\midrule
BERT   & 54.9     &  \textbf{77.2} & 64.2    &  52.7 & 78.8           &  63.2 \\
\MODEL  & \ \ \textbf{58.6}\starr       &  75.4 & \ \ \textbf{65.9}\starr       & \ \  \textbf{55.2}\starr & \ \ \textbf{80.7}\starr       &  \ \ \textbf{65.6}\starr \\
\midrule
F1 $\Delta$\%  & & & +2.6\% & & & +3.8\% \\
\bottomrule
\end{tabular}
}
\caption{
Main settings using the errors from the \video ASR model.
Format is similar to \Cref{tab:error-detection}.
}
\label{tab:another_asr}
\vspace{-0.5cm}
\end{table}

    \section{Related Work}
\label{sec:related-work}

\paragraph{ASR Confidence Scores} 
are used to evaluate reliability of recognition results \cite{CONFIDENCE_SCORES_SURVEY}. In modern ASR models, a separate confidence network is usually trained using a held-out dataset \cite{CONFIDENCE_SCORE_RESEARCH_1, CONFIDENCE_SCORE_RESEARCH_3}.

\paragraph{Uncertainty Calibration}
adapts a models prediction probabilities to better reflect their true correctness likelihood \cite{CALIBRATION_1}.
We provide the Brier Scores (evaluating calibration) for our dataset in \Cref{tab:brier-scores}.
AED models, which perform a binary classification -  \ERR or \NOTERR, do not explicitly use calibration. For example in \CONFIDENCE, \BERTorC and \BERTandC we tune the threshold to an optimal value, and since most calibration techniques will preserve the relative scores ordering, better calibration will not improve performance. \BERTC and \MODEL  do not rely on calibrated scores, since deep neural networks can model non linear relationships \cite{NN}.

\paragraph{\AED.}
We provide a brief summary of relevant AED papers, for a more thorough review of AED we refer the reader to \citet{ERROR_CORRECTION_REVIEW}. 

\citet{ERROR_DETECTION_WITH_CONFIDENCE} used data mining models, leveraging features from confidence scores and a linguistics parser. 
\citet{ERROR_DETECTION_1} used logistic regression with features extracted from
confusion networks. \citet{ERROR_DETECTION_2} used a Markov Chains classifier. \citet{ERROR_DETECTION_3} focused on spoken translation using confidence from a machine translation model, posteriors from entity detector and a word boundary detector. 
 
Modern Transformer-based approaches have not addressed the \AED task directly. 
A few attempts were made to apply Transformers for ASR \emph{error correction}, using a sequence-to-sequence models to map directly between the ASR hypothesis and the correct (reference) transcription \cite{PREV_SEQ2SEQ_ERROR_CORRECTION_1, PREV_SEQ2SEQ_ERROR_CORRECTION_2, FASTCORRECT2, FASTCORRECT}. To the best of our knowledge, our work is the first to address \AED using the Transformer and to introduce representation for ASR confidence scores in a Transformer-based ASR post-processing model.

\begin{table}[t]
\small
\centering
\scalebox{0.8}{
\begin{tabular}{l|ccc|ccc}
\toprule
& \multicolumn{3}{c|}{\other \ $\rightarrow$ \clean} & \multicolumn{3}{c}{\clean \ $\rightarrow$ \other} \\
  & R & P  & F1   & R &  P & F1 \\
\midrule
BERT    &    61.2           & 73.5 & 66.8              & 42.9          &  \textbf{82.2}             & 56.4 \\
\MODEL & \ \ \textbf{62.8}\starr  &   \ \ \textbf{75.8}\starr   & \ \ \textbf{68.7}\starr & \ \ \textbf{47.7}\starr     & \ \ 79.8\starr &  \ \ \textbf{59.7}\starr    \\
\midrule
F1 $\Delta$\%  & & & +2.8\% & & & +5.9\% \\
\bottomrule
\end{tabular}
}
\caption{
Robustness settings using the errors from the \video ASR model. Format is similar to \Cref{tab:error-detection-robustness}.
}
\label{tab:error-detection-robustness-video}
\vspace{-0.5cm}
\end{table}

    \section{Conclusion}
\label{sec:conclusion}
We introduced \ace \ \MODEL, an approach for embedding ASR word-level confidence scores into a Transformer-based ASR error detector. \MODEL jointly encodes the scores and the transcription hypothesis into a contextualized representation. 

Our experiments demonstrated that the ASR word-level confidence scores are useful on top of the transcription hypothesis text, yet it is not trivial to effectivelly combine these signals. We showed that performing such combination using RED-ACE leads to significant performance gains, as well as increased robustness to changes in the audio quality, which can be crucial for real-world applications. 

In addition, we published a novel AED dataset that allows researchers to train and evaluate AED models, without the need to run ASR models. It also ensures the full reproducibility of our results in case Google Cloud models will change over time.

In future work, we would like to leverage additional signals from the ASR model (such as alternative hypotheses), as well as explore the benefits of confidence scores for \emph{error correction} models.
    \section{Limitations}

\paragraph{Limitations} 
Our approach does not account for ASR errors where the ASR system simply deletes output words. However, it is not clear whether those cases are of a practical use for an AED application that highlights incorrect words in the hypothesis, as in this case there is nothing to highlight.
More specifically, our approach does not consider \emph{isolated} deletions. 

To illustrate that, let’s first consider an example in which 2 words were transcribed as 1 word, meaning that 1 word was omitted in the transcription. For example, if \emph{"a \underline{very big} cat"} was transcribed as \emph{"a \underline{small} cat"}. An \AED application would ideally highlight the word \emph{"\underline{small}"} as a transcription error.
This case is actually covered by our approach, even though one word is omitted in the transcription, because when creating the \AED dataset we will label “small” as an error and train the model accordingly (details in \cref{sec:dataset}). 

The cases that are not covered are when the ASR model omits words while all the surrounding words are transcribed correctly. For example \emph{"a \underline{very} big cat"} that was transcribed as \emph{"a big cat"}. In this case, all the words in the transcription hypothesis are correct words and our approach is not expected to discover any error. We chose not to cover those cases as it is not clear if they are useful for an error detection application, that usually needs to highlight incorrect words in the hypothesis. In addition, ignoring those cases is also in-line with previous work \cite{ERROR_DETECTION_WITH_CONFIDENCE}. Finally, our analysis showed that those cases are extremely rare, for example in \clean they occur only in $0.37\%$ of the words.

\paragraph{Risks} A possible risk posed by an AED system could be caused by an over-reliance on it. Whereas without AED, the entire output of an ASR system may have been manually verified, with AED only parts of output which the AED flagged may be verified, leading to errors remaining that were not found by the AED system. 

\section*{Acknowledgements}
We thank the reviewers for their valuable suggestions for improving this work. We would also like to thank Gal Elidan, Idan Szpektor, Eric Malmi, Yochai Blau, Amir Feder, Andrew Rosenberg, Françoise Beaufays and Avinatan Hassidim for reviewing the paper and providing useful feedback.
    
    \bibliography{anthology,custom}
    \bibliographystyle{acl_natbib}
    
    \clearpage

\appendix

\section{Appendix}
\label{sec:appendix}

\subsection{Implementation Details}
\label{impl:details}
\paragraph{Training.} We fine-tune our BERT-based \cite{BERT} model with a batch size of 512\footnote{We choose the best among 128, 512 and 1024, based on tagging accuracy on the development set.}, a weight decay of 0.01, and a learning rate of 3e-5\footnote{We choose the best among 5e-5, 4e-5, 3e-5, and 2e-5, based on tagging accuracy on the development set.}. The maximum input length is set to 128 tokens. We pad shorter sequences and truncate longer ones to the maximum input length. We use the cross-entropy loss function, optimizing the parameters with the AdamW optimizer. 
We train for a maximum of  500 epochs and choose the checkpoint with the maximum tagging accuracy on the development set.\footnote{For \MODEL the tagging accuracy was 95.4 on \clean and 89.7 on \other.} The best checkpoint was found at epochs 100-150 after approximately 8 hours of training time. All models were trained on  TPUs (4x4). BERT-base has 110 million parameters, the inclusion of confidences embeddings for \MODEL added ~10k additional parameters. The confidence embedding matrix is randomly initialized with truncated normal distribution\footnote{\url{https://www.tensorflow.org/api_docs/python/tf/keras/initializers/TruncatedNormal}}. 

If a single word is split into several tokens during BERT's tokenization, all the corresponding tokens get the confidence score of the original word. To predict word-level errors (used throughout the paper), we treat a word as an error if one of its tokens was tagged as error by the model. To predict span-level errors (reported for completeness in \Cref{tab:span-level}), we treat every sequence of errors as one error-span and every sequence of correct words as a correct-span.

\paragraph{Binning.}
\label{sec:effect_of_binning}
Table \ref{tab:binning} presents results for different binning algorithms and bin sizes. 
For binning algorithms we use: (1) simple equal-width binning and (2) quantile-based discretization (equal-sized buckets). 
We note that there is no significant difference between the results. In our main experiments we used equal width binning with 10 bins. For special tokens,\footnote{[CLS] and [SEP] in case of BERT.} that do not have confidence scores, we chose to allocate a dedicated bin.

\paragraph{Statistics Significance Test.} 
In \cref{tab:error-detection}, in addition to the main results, we provide a statistic significance tests results. For this purpose we pseudo-randomly shuffle all words in our test set, split them up into 100 approximately equally sized subsets, and compute recall, precision and F1 for each of them for the baseline and \MODEL models. We then apply the Student’s paired t-test with $p < 0.05$ to these sets of metrics. To determine statistical significance in F1 $\Delta$\% between different setups evaluated on the same data set, F1 $\Delta$\% is computed for each of the given subsets, and the same significance test is applied to the resulting sets of F1 $\Delta$\% between two setups. 

\begin{table}[t]
\small
\centering
\begin{tabular}{l|l|cccc}
\toprule
Binning algorithm & \# Bins  & R & P & F1   \\ 
\midrule
\multirow{3}{*}{Equal width bins}   & 10       &   \textbf{64.1} &   79.9 & \textbf{71.1}        \\
                                    & 100       &   62.5 &   80.5 &   70.4        \\
                                    & 1000       &   63.2 &   80.7 &  70.9        \\
\midrule
Equal size bins   & 10       &   63.0 &   \textbf{81.5} &   \textbf{71.1}        \\
\bottomrule
\end{tabular}
\caption{
Effect on different binning strategies (\other).
}
\label{tab:binning}
\end{table}

\begin{table}[t]
\small
\centering
\begin{tabular}{l|l|r|r}
\toprule
Pool & Subset Name & Audio Hours & \# Examples   \\ 
\midrule
\multirow{4}{*}{Clean} &    \emph{train-clean-100} & 100.6  & 28,539      \\
    & \emph{train-clean-360} & 363.6   & 104,014     \\
    & \emph{dev-clean} & 5.4    & 2,703    \\
    & \emph{test-clean} & 5.4   & 2,620     \\
\midrule
\multirow{3}{*}{Other}    & \emph{train-other-500} & 496.7    & 148,688    \\
    & \emph{dev-other} & 5.3   & 2,864     \\
    & \emph{test-other} & 5.1  & 2,939      \\
\bottomrule
\end{tabular}
\caption{
LibriSpeech corpus subsets statistics.
}
\label{tab:librispeech_subsets}
\vspace{-0.2cm}
\end{table}

\subsection{Published \AED Dataset}
\label{appendix:published_dataset}
As described in \S \ref{sec:dataset}, we generate our own \AED dataset. 
To this end we transcribe the LibriSpeech corpus using 2 modes from Google Cloud Speech-to-Text API.\footnote{\url{https://cloud.google.com/speech-to-text}} We choose the \default model as our main model and the \video model as the additional model\footnote{\url{https://cloud.google.com/speech-to-text/docs/basics\#select-model}}. We also enable the word-level confidence in the API.\footnote{\url{https://cloud.google.com/speech-to-text/docs/word-confidence\#word-level_confidence}} 
Our submission includes the \AED dataset as well as the predictions of our models on the test sets. We hope that our dataset will help future researchers and encourage them to work on \AED.

\paragraph{The LibriSpeech Corpus Details.}
\label{appendix:libriSpeech_corpus}
We provide here additional details abut the LibriSpeech corpus.\footnote{\url{https://www.openslr.org/12/}} 
The corpus contains approximately 1000 hours of English speech from read audio books.
The corpus contains \emph{clean} and \emph{other} pools.
The training data is split into three subsets: \emph{train-clean-100}, \emph{train-clean-360} and \emph{train-other-500}, with approximate sizes of 100, 360 and 500 hours respectively. Each pool contains also a development and test sets with approximately 5 hours of audio. Full data split details can be seen in \cref{tab:librispeech_subsets}. We note that the \#Examples is slightly different than the numbers in our dataset (see \cref{tab:dataset-stats}). When transcribing with Google Cloud API, we occasionally reached a quota limit and a negligible number of examples was not transcribed successfully (up to 2\% per split). The \clean pool contains 2 training sets, we used the larger one in our dataset (\emph{train-clean-360}).

\begin{figure}[t]
 \centering
    \includegraphics[width=0.91\columnwidth]{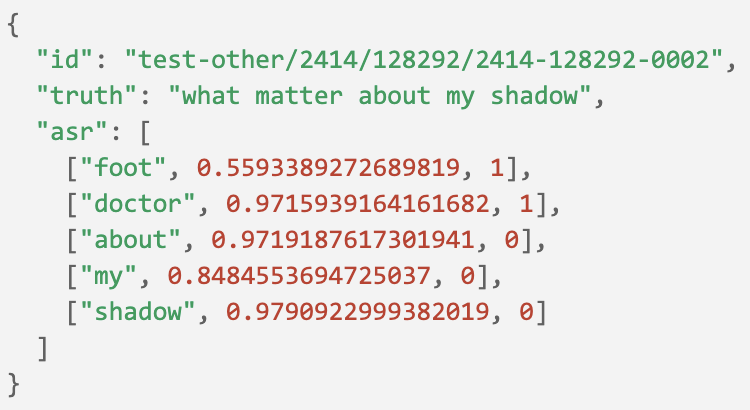}
    \caption{
    A single example from our AED dataset.
    }
    \label{fig:dataset_example} 
\end{figure}

\paragraph{Annotation Description.}
A single example from our \AED dataset can be seen is \cref{fig:dataset_example}. The annotation contains the ASR hypothesis words, the corresponding word-level confidence scores and the \ERR or \NOTERR label.

\paragraph{License.}
This data as well as the underlying LibrSpeech ASR corpus are licensed under a Creative Commons Attribution 4.0 International License\footnote{\url{http://creativecommons.org/licenses/by/4.0/}}.

\begin{table}[t]
\small
\centering
\begin{tabular}{ccl}
\toprule
\multicolumn{1}{c}{ASR Model} & \multicolumn{1}{c}{Pool} & \multicolumn{1}{c}{Brier Score} \\
\midrule
\multirow{2}{*}{\emph{default}} & \multirow{1}{*}{\clean} &  0.069  \\
\cmidrule{2-3}
& \multirow{1}{*}{\other}  & 0.142     \\
\midrule
\multirow{2}{*}{\emph{video}} & \multirow{1}{*}{\clean} &  0.06      \\
\cmidrule{2-3}
& \multirow{1}{*}{\other}  & 0.1     \\
\bottomrule
\end{tabular}
\caption{ 
Brier Scores (evaluating confidence scores calibration, lower is better) for our dataset.
}
\label{tab:brier-scores}
\end{table}

\makeatletter
\setlength{\@fptop}{0pt}
\makeatother
\begin{table}[t]
\small
\centering
\scalebox{0.83}{
\begin{tabular}{l|ccc|ccc}
\toprule
& \multicolumn{3}{c|}{   \clean} & \multicolumn{3}{c}{   \other} \\
  & R & P  & F1   & R &  P & F1 \\
\midrule
\CONFIDENCE    & 31.0 & 27.5 & 29.1 & 23.4 & 20.2 & 21.7  \\
\MLM       &    27.4   &    12.3   &    17.0   &    22.2   &    11.6&    15.2     \\
\BERT       &    47.1          & 59.6 & 52.6   &   37.1 & 46.6 & 41.3  \\
\BERTandC    & 44.5 & 55.5 & 49.4 & 35.5 & 43.7 & 39.2 \\
\BERTorC    & 48.9 & 51.0 & 49.9 & 40.6 & 40.2 & 40.4 \\
\BERTC   & 45.4 & 59.9 & 51.7 & 37.5 & 48.0 & 42.1 \\
\midrule
\MODEL  &  \textbf{49.4}  & \ \ \textbf{63.6}\starr & \ \ \textbf{55.6}\starr     &  \ \ \textbf{42.1}\starr &  \ \ \textbf{50.9}\starr  & \ \ \textbf{46.1}\starr  \\
\midrule
F1 $\Delta$\%  & & & +5.4\% & & & +9.5\% \\
\bottomrule
\end{tabular}
}
\caption{
Span-level results for the main settings using the errors from the \default ASR model. The format is similar to \Cref{tab:error-detection}.
}
\label{tab:span-level}
\end{table}

\end{document}